\documentclass[final]{cvpr}

\usepackage{times}
\usepackage{epsfig}
\usepackage{graphicx}
\usepackage{amsmath}
\usepackage{amssymb}
\usepackage[font=small]{caption}
\usepackage{subcaption,booktabs}

\newcommand\blfootnote[1]{%
  \begingroup
  \renewcommand\thefootnote{}\footnote{#1}%
  \addtocounter{footnote}{-1}%
  \endgroup
}

\usepackage[pagebackref=true,breaklinks=true,colorlinks,bookmarks=false]{hyperref}

\makeatletter
\DeclareRobustCommand\onedot{\futurelet\@let@token\@onedot}
\def\@onedot{\ifx\@let@token.\else.\null\fi\xspace}
 
\def\ie{\emph{i.e}\onedot}

\makeatother

\def\cvprPaperID{****} 
\def\confYear{CVPR 2021}

\begin{document}

\title{Feature Combination Meets Attention: Baidu Soccer Embeddings and Transformer based Temporal Detection}

\author{Xin Zhou* \ \ \ \ \ \ \ \ Le Kang* \ \ \ \ \ \ \ \ Zhiyu Cheng* 
\and
Bo He \ \ \ \ \ \ \ \ Jingyu Xin
\and
Baidu Research\\
1195 Bordeaux Dr, Sunnyvale, CA 94089 USA\\
{\tt\small \{zhouxin16, kangle01, zhiyucheng, v\_hebo02, xinjingyu\}@baidu.com}
}

\maketitle

\begin{abstract}
    With rapidly evolving internet technologies and emerging tools, sports related videos generated online are increasing at an unprecedentedly fast pace. To automate sports video editing/highlight generation process, a key task is to precisely recognize and locate the events in the long untrimmed videos. In this tech report, we present a two-stage paradigm to detect what and when events happen in soccer broadcast videos. Specifically, we fine-tune multiple action recognition models on soccer data to extract high-level semantic features, and design a transformer based temporal detection module to locate the target events. This approach achieved the state-of-the-art performance in both two tasks, i.e., action spotting and replay grounding, in the SoccerNet-v2 Challenge, under CVPR 2021 ActivityNet workshop. Our soccer embedding features are released at https://github.com/baidu-research/vidpress-sports. By sharing these features with the broader community, we hope to accelerate the research into soccer video understanding.

\end{abstract}


\section{Introduction}
\begin{figure}
    \centering
    \includegraphics[width=6.8in]{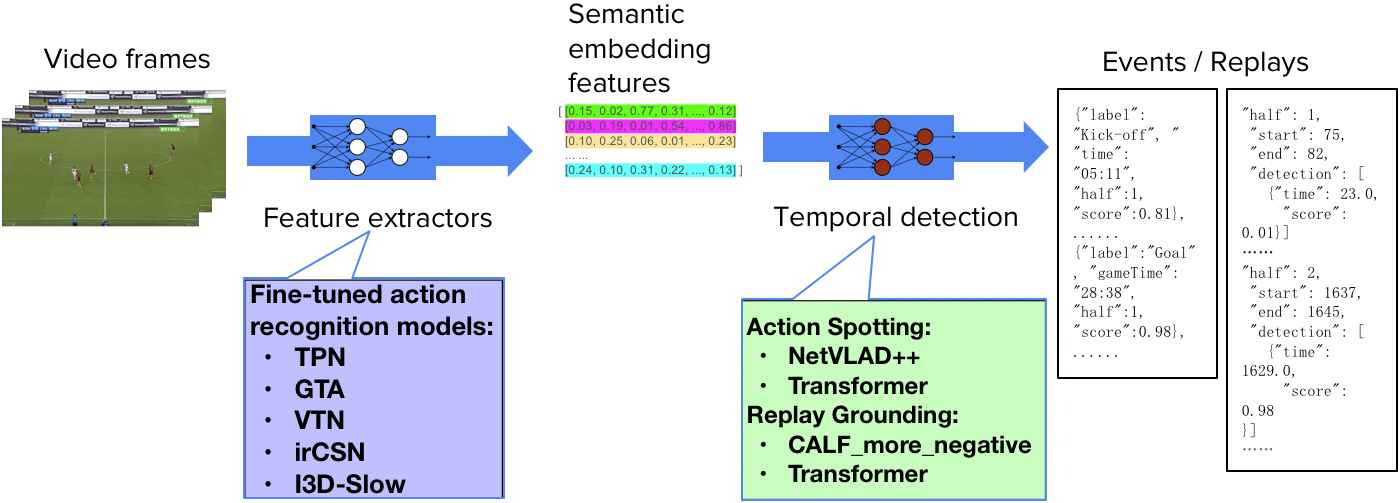}
    \caption{Our two-stage paradigm for action spotting and replay grounding in soccer videos}
    \label{fig:pipeline}
\end{figure}

\blfootnote{ * denotes equal contributions. The order was determined using Python’s random.shuffle()}
Creating sports highlight videos often involves human efforts to manually edit the original untrimmed videos. The most popular sports videos often comprise short clips of a few seconds, while for machines to understand the video and spot key events precisely is very challenging. In this tech report, we present a two-stage paradigm (see Figure.\ref{fig:pipeline}) to solve two problems in understanding soccer videos and detecting target events: action spotting and replay grounding, which are defined in SoccerNet-v2 challenge \cite{Delige2020SoccerNetv2A}. The action spotting task aims at spotting actions such as goal, shots-on target, shots-off target, yellow card, red card, etc, in a complete video of soccer game. The replay grounding task is to ground the timestamps of the actions represented in a specific replay. In our approach, the same first stage is shared in both two tasks, which uses fine-tuned action recognition models to extract semantic features, and the second stage consists of a temporal detection module tailored for each task.

\subsection{Related work}
In sports analytics, many computer vision technologies are developed to understand sports broadcasts \cite{THOMAS20173}. Specifically in soccer, researchers propose algorithms to detect players on field in real time \cite{Cioppa_2020_CVPR_Workshops}, analyze pass feasibility using player’s body orientation \cite{Sangesa2020}, incorporate both audio and video streams to detect events \cite{Vanderplaetse2020}, recognize group activities on the field using broadcast stream and trajectory data \cite{Sanford2020}, aggregate deep frame features to spot major game events \cite{Giancola_2018_CVPR_Workshops}, and leverage the temporal context information around the actions to handle the intrinsic temporal patterns representing these actions \cite{Cioppa2020Context,Giancola_2021_CVPR_Workshops}. 

\subsection{Contributions}



In this tech report, our main contributions can be summarized as the following:


$\bullet$ Taking advantage of multiple recent action recognition models pretrained on large-scale video datasets, we extract semantic features of soccer videos by fine-tuning each model as the feature extractor, on an auxiliary snippet dataset which is derived from the original SoccerNet-v2 dataset. We concatenate and normalize the features obtained from each model to generate Baidu soccer embeddings. The proposed feature combination significantly improves both action spotting and replay grounding performance.

$\bullet$ We propose a novel transformer based temporal detection module which achieves the state-of-the-art performance in both action spotting task and replay grounding task in the SoccerNet-v2 Challenge, under 2021 CVPR ActivityNet Workshop.



\section{Feature Extraction} \label{FeatExtract}

Both the previous competitive method NetVLAD++ \cite{Giancola_2021_CVPR_Workshops} for action spotting and the baseline method $CALF\_more\_negative$ (Cmn) \cite{Delige2020SoccerNetv2A} for replay grounding use per-frame features extracted by ResNet~\cite{He2016DeepRL} pretrained on ImageNet~\cite{krizhevsky2012imagenet}. 
However, we believe that features that are tailored for the soccer broadcast videos can improve the performance of the spotting module. 
We fine-tune multiple action recognition models on snippets of SoccerNet-v2 videos, and in the test stage we also extract features from videos (clips of frames), rather than on a per-frame basis. 
We fine-tune multiple action recognition models on the task of action classification. The models we use include TPN \cite{Yang2020TemporalPN}, GTA \cite{he2020gta}, VTN \cite{neimark2021video}, irCSN \cite{Tran2019}, and I3D-Slow \cite{Feichtenhofer2019SlowFastNF}. In order to perform such fine-tuning, we construct an 18-class snippet dataset by extracting snippets, each with 5 seconds long, from all the videos. Each snippet centers at one of the 17 classes of events or randomly samples from background (non-event). We apply each fine-tuned action recognition model on the temporal sliding windows of videos, and concatenate output features along the feature dimension.


Here we briefly introduce the five pretrained action recognition models we choose to fine-tune on soccer data. The temporal pyramid network (TPN) \cite{Yang2020TemporalPN} efficiently incorporate visual information of different speeds at feature level, and can be integrated into 2D or 3D backbone networks. It can achieve $78.9\%$ top-1 accuracy on Kinetics-400 dataset with a ResNet-101 backbone network. The global temporal attention (GTA)   mechanism for video action classification proposed in  \cite{he2020gta} models global spatial and temporal interactions in a decoupled manner. It captures temporal relationships on both pixels and semantically similar regions. On Kinetics-400 dataset, GTA achieves $79.8\%$ top-1 accuracy when applied on SlowFast-R101 network. The video transformer network (VTN) \cite{neimark2021video} adopts transformer based network structure for video understanding. It trains faster than 3D CNN networks and can achieve $78.6\%$ top-1 accuracy on Kinetics-400 dataset. The interaction-reduced channel-separated convolutional networks (irCSN) introduced in  \cite{Tran2019} factorizes 3D convolutions by separating channel interactions and spatio-temporal interactions. In this way, a regularization effect is observed when training the network. The authors also pretrained this network on a large-scale dataset, IG-65M \cite{Ghadiyaram2019LargeScaleWP}, before fine-tuning on Kinetics-400 where achieved $82.6\%$ top-1 accuracy. The I3D-Slow network preserves the slow pathway, which operates at low frame rate and captures spatial semantics in the SlowFast framework   \cite{Feichtenhofer2019SlowFastNF}. It is pretrained with OmniSource \cite{duan2020omni} data and can reach $80.4\%$ top-1 accuracy on Kinetics-400.

\section{Temporal Detection}

\begin{figure}
    \centering
    \includegraphics[width=5.8in]{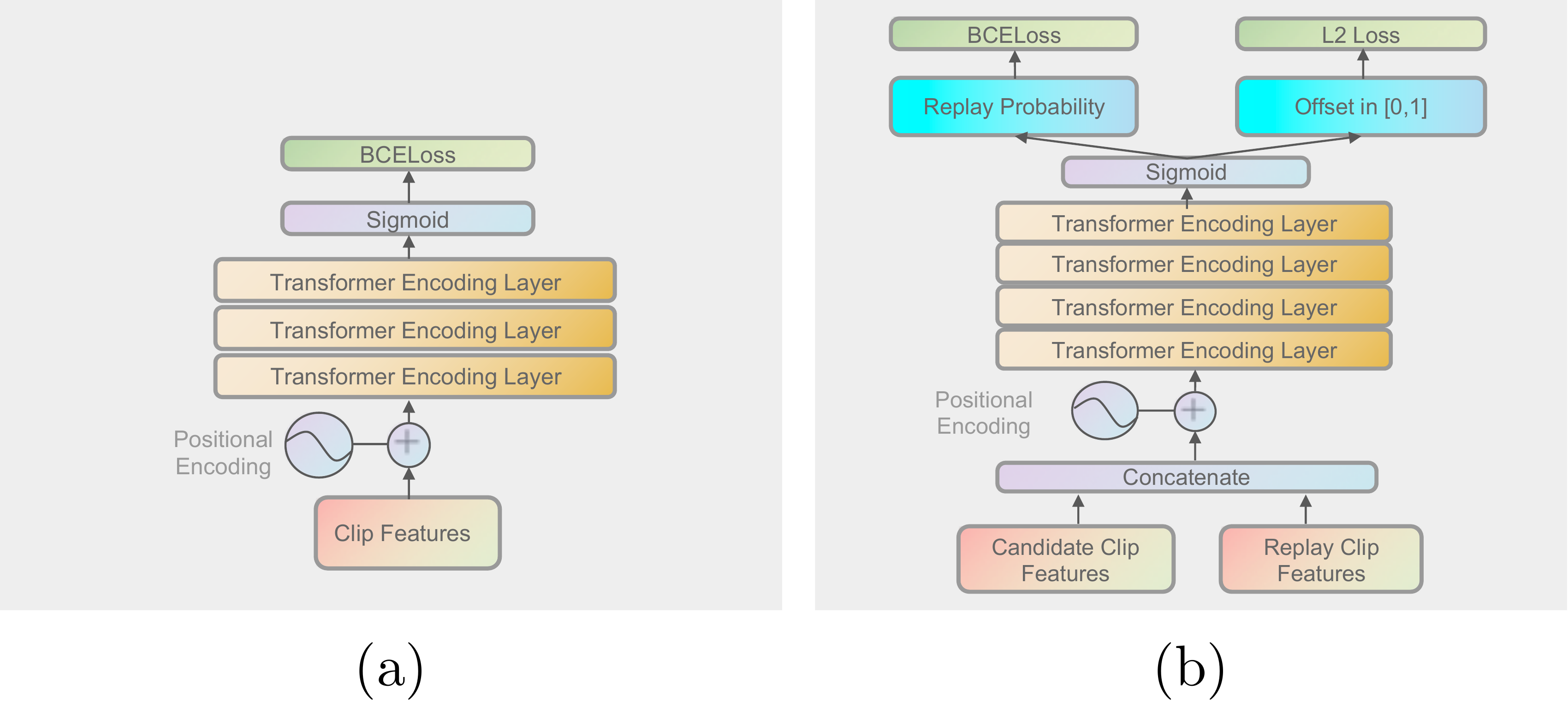}
    \caption{Our transformer based models for (a) action spotting and (b) replay grounding}
    \label{fig:transformer}
\end{figure}

In this section, we present the temporal detection module in our two-stage paradigm for soccer video understanding. Specifically, 1) NetVLAD++ and transformer for action spotting task, and 2) Cmn and transformer for replay grounding task.

\subsection{Action Spotting}
Given the combined features described in Section \ref{FeatExtract} as input, a NetVLAD++ \cite{Giancola_2021_CVPR_Workshops} model can yield much higher performances than the original ResNet features. We also implemented other methods including 1D-ResNet \cite{He2016DeepRL} and Transformer \cite{vaswani2017attention}, and they can achieve similar results. Since the Transformer obtains the best performance on the challenge set, we describe its implementation details as follows.

For the transformer model, as in \cite{vaswani2017attention}, we use sine, cosine positional encoding. We only use the encoder part of the model to get an output of dimension 18 to represent the 18-class probabilities. As shown in Figure \ref{fig:transformer}(a), we create three transformer encoding layers after the positional encoding. We choose 4 heads and hidden dimension of 64 for the encoding layers. In training, we adopt mix-up augmentation \cite{zhang2017mixup} to reduce over-fitting.

To further improve the performance, we make the following adjustments: (a) train the video recognition models for feature extraction on an enlarged dataset, which is the aggregation of train, valid, and test sets in the snippet dataset we mentioned in Section \ref{FeatExtract}, together with snippets extracted from extra 77 games we collected in Spain Laliga 2019-2021 seasons; (b) train the spotting module on the aggregation of train, valid, and test sets; (c) change hyper parameters including feature dimension size, batch size, learning rate, chunk size and NMS window size. Details will be elaborated in the experiment section.

\subsection{Replay Grounding}

In this section, we first analyze replay annotations of 500 games in the SoccerNet-v2 dataset, and then discuss the baseline grounding module Cmn and our transformer based grounding module. 

\subsubsection{Replay pattern analysis}

\begin{figure}
    \centering
    \includegraphics[width=6.8in]{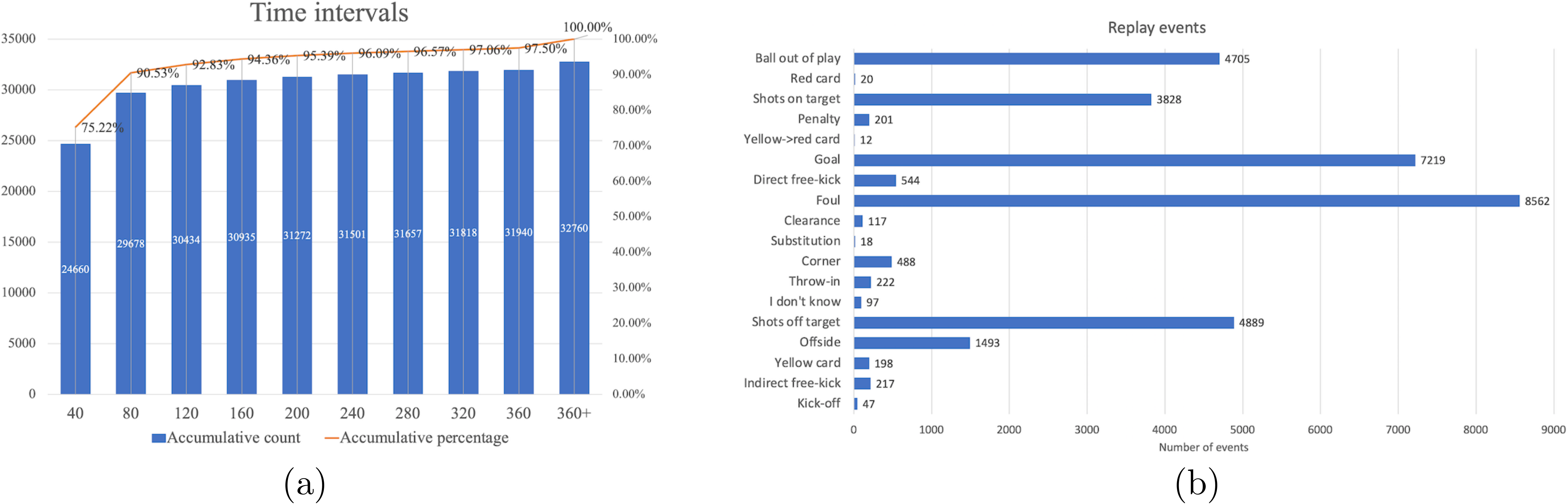}
    \caption{Replay pattern analysis. (a) Time intervals between the end timestamp of a replay and the original event's timestamp. (b) Replay events.}
    \label{fig:replay}
\end{figure}

To better choose hyperparameters for our model and improve grounding results, we analyze all replay annotations. Figure \ref{fig:replay}(a) shows the distributions of time intervals between the end timestamp of a replay and the original event's timestamp. We found that $92.83\%$ of the time intervals fall in the range of $0\sim 120$ seconds. Therefore, for efficient training and inference, we design the transformer based grounding module and design filtering strategies in post processing for Cmn module to focus in this range. Figure \ref{fig:replay}(b) shows the number of different types of events in ground-truth replay annotations. We found that the top 3 events in terms of total counts are foul, goal, and shots-off target respectively. This observation helps us design fusion strategies in post processing which will be described in the experiments session.  


\subsubsection{Transformer based grounding module}
\label{replay:transformer}

To capture the relationship between the replay clip and candidate clip, we apply a transformer encoder. Following the configurations in \cite{vaswani2017attention}, we choose sine, cosine positional encoding. As shown in Figure \ref{fig:transformer}(b), the input consists of semantic features of a candidate clip and a replay clip. We stack up 4 encoding layers and get an output with 2 dimensions (replay probability and positional offset) which align with the spotting output dimensions in the baseline approach in Cmn from \cite{Delige2020SoccerNetv2A}. Unlike the baseline grounding module Cmn, we disabled segmentation loss. We apply the binary cross-entropy loss (BCELoss) to train the replay probability and L2 loss to train the positional offset.


Since our fine-tuned features work better with shorter clips (feature extractors trained on 5-second snippets) and to prevent the module from over-fitting, we adjust the video chunk size to $30$ seconds. We also only fine-tune the grounding module on video chunks extracted from at most $120$ seconds before the start of replays since the feature extractors are already trained on all event snippets from full match videos, and most replays happen within the $120$ seconds after the original events according to our replay pattern analysis. In the $120$ seconds clip, $4$ positive and $4$ negative samples are given to the transformer such that we have sufficient data to better learn the grounding component of the output.






\section{Experiments}

\begin{table}[t]
    \vspace{-0.1in}
    \small
    \centering
    \caption{Video action recognition models for extracting semantic features}
    \begin{tabular}{@{}c|c|c|c@{}} 
        \toprule
        Arch                                                    & Backbone      & Dim   & Pretrain\\
        \midrule
        TPN \cite{Yang2020TemporalPN}               & ResNet50/101  & 2048  & K400\\ 
        GTA \cite{he2020gta}                        & ResNet50      & 2048  & K400\\ 
        VTN \cite{neimark2021video}                 & ViT-Base      & 384   & K400 \\ 
        irCSN \cite{Tran2019}     & ResNet152     & 2048  & IG65M + K400\\ 
        I3D-Slow \cite{Feichtenhofer2019SlowFastNF} & ResNet101     & 2048  & OmniSource\\ 
        \bottomrule
    \end{tabular}
    \label{table:1}
    \vspace{-0.1in}
\end{table}

\subsection{Dataset and Evaluation}\label{dataeval}
The SoccerNet-v2 dataset contains broadcast videos of 550 soccer games. We mainly use the LQ version of these videos at 25fps with a resolution of $398\times 224$. In addition, we collect broadcast videos of 77 extra games in Spain Laliga 2019-2021 seasons. We check the extra videos and guarantee that they do not contain any game from the SoccerNet-v2 challenge set. We annotate these videos in the similar protocol as SoccerNet-v2, and convert videos into LQ in order to fine-tune feature extractors.

We report the performance of our methods using the Average-mAP metric introduced by SoccerNet-v2.

\subsection{Implementation Details} 
For the feature extraction stage, Table \ref{table:1} shows all the action recognition models we use with their main configurations. 
These models are pretrained from various large-scale datasets, including Kinetics-400 (K400)\cite{46330}, IG65M \cite{Ghadiyaram2019LargeScaleWP}, and Omnisource\cite{duan2020omni}. All models are fine-tuned on SoccerNet-v2 snippets to reach a reasonable top-1 classification accuracy between $78\%$ and $82\%$ (1-view test) on the test set. At test time, all models slide on the videos and produce features at 1fps. 

To boost the performance on the challenge set, we also fine-tune the feature extractors (action recognition models) on an enlarged snippet dataset, which contains snippets from the train, valid, test videos of SoccerNet-v2 and videos of 77 extra games. We denote the produced features as mega features if the extractors are fine-tuned on the enlarged snippet dataset.


In our experiments, the spotting or grounding module is trained in two modes, regular and ultra. In the regular mode, train, valid, and test set each serves its own purpose following the same setting as the reference NetVLAD++ \cite{Giancola_2021_CVPR_Workshops} method for action spotting or Cmn \cite{Delige2020SoccerNetv2A} for replay grounding. In the ultra mode, we make the spotting/grounding module to learn from as much data as possible, thus we use all features from train, valid, and test sets to train the spotting/grounding module, for a fixed amount of epochs. 


For the action spotting task, we use a learning rate of $10^{-4}$ for the NetVLAD++ model. The ultra mode training stops at $40$ epochs. For the tranformer model, we use a learning rate of $5\times10^{-4}$ and stop at $50$ epochs.
For the replay grounding task, a learning rate of $2\times10^{-4}$ is adopted using the transformer model and training stops at $40$ epochs in ultra mode.



\subsection{Results and Analysis}
\subsubsection{Action Spotting}
Table \ref{table:2} shows the performance of our methods with different configurations. When using only one (ordinary) feature from TPN-r50 and perform spotting using NetVLAD++ in the regular mode, as shown in the first row of Table \ref{table:2}, we achieve an Average-mAP of $62.96\%$ and $62.35\%$ on the test and challenge set, respectively, which is about $9\%\sim10\%$ gain over than the reference NetVLAD++'s $53.30\%$ and $52.54\%$ on test and challenge sets, respectively. This result shows the superior advantage of using a recent action recognition model fine-tuned on SoccerNet-v2 as a feature extractor. 

When using 4 features or 5 features combined as the input of the spotting module, as shown in row 3 and 6 in Table \ref{table:2}, we obtain about $5\% \sim 9\%$ gain on both the test and the challenge sets over the 1 feature case. Such comparison shows feature combination also significantly improves the performance.

Training the spotting module in the ultra mode with 4 mega features results in a challenge Averge-mAP of $68.68\%$ (row 5 in Table \ref{table:2}), compared to the regular mode with the ordinary features at $67.51\%$ (row 3 in Table \ref{table:2}) and the regular mode with the mega features at $67.57\%$ (row 4 in Table \ref{table:2}). This comparison indicates that it improves the generalization power only if both stages use more data for training.

Comparing row 6 and 8 in Table \ref{table:2}, we can see that adjusting chunk size/NMS window size from $15/30$ to $7/20$ leads to additional $1.5\% \sim 2\%$ gain on the test and challenge sets. 

Our transformer based spotting module, trained in the ultra mode with mega features plus adjusted chunk size/NMS window size, achieves the best challenge Average-mAP at $74.84\%$ as shown in row 10 in Table \ref{table:2}. While the NetVLAD++ based module, trained in the same setting, achieves similar performance: $74.63\%$ on the challenge set.

\subsubsection{Replay grounding}

We summarize our experimental results for replay grounding task in Table \ref{table:results}. As we can see from the table, taking the fine-tuned features as input significantly improved the grounding results compared to the baseline average-AP in Row 1. In addition, based on the same grounding module, combining more features extracted with different action recognition models leads to further improvements. We also observed superior performance by using a large batch size of $128$. 


To further improve the performance, we also investigated several post-processing techniques to refine the grounding module outputs, based on our analysis of the replay pattern:

$\bullet$ Filtering: For Cmn based grounding module, we eliminate all spotting results 1) after the end timestamp of the replay, 2) more than 100 seconds (threshold) prior to the end timestamp of the replay. Note the filtering threshold in Row 9 was set to 120 seconds.

$\bullet$ Fusion: Taking advantage of the action spotting results, we incorporate the action classification information into the replay grounding task.
For each queried replay clip with start time $T$, we adopt the following procedures. First, we filter spotting predictions with top-3 most frequent labels of replay actions (\ie, 'Foul', 'Goal', 'Shots-off target') and with the predicted scores higher than $S$. Second, the first and second nearest spotting predictions to the replay clip start time $T$ are selected, and satisfying the constraint that each prediction falls into the temporal window range $[T-W, T]$, because the actual action should happen before the relay action. Third, we use the spotting confidence score as the replay grounding confidence score, and the score of the nearest prediction is multiplied with a factor $\beta_1$ and the second-nearest prediction is multiplied with $\beta_2$. Through experiments, we find that $W=42, S=0.02, \beta_1=1.25, \beta_2=0.8$ achieves the best performance.

$\bullet$ NMS: We combine the grounding results from Cmn model and transformer model, normalize all event scores to $[0.0, 1.0$, and apply an NMS (Non Maximum Suppression) to reduce positive spots within a window size of $25$. 

The combined post processing techniques achieved a decent performance improvement, around $12\%$ comparing Row 5 and Row 8. However, our best result is achieved using the transformer based grounding module described in Section \ref{replay:transformer} and trained in ultra mode, which is $71.9\%$ as shown in Row 9. Specifically, we trained the transformer in $40$ epochs and it took about 3 hours on a TitanX GPU, which is significantly faster than training the Siamese neural network in the baseline approach.

\section{Conclusion}
We present the two-stage paradigm for the action spotting and replay grounding tasks, and fine-tune action recognition models on soccer videos to extract semantic features and using Transformer in the spotting and grounding modules. We achieve the state-of-the-art results on the challenge set of SoccerNet-v2. Developing the proposed action spotting and replay grounding pipeline in soccer videos is the first step for machines to fully understand sports videos, which can dramatically benefit sports media, broadcasters, YouTubers or other short video creators. We believe the presented methodology can be extended to detect and locate events in other sports videos. We released our semantic features to support further soccer video understanding research.

\begin{table}[t]
    \small
    \centering
    \caption{Experimental results using different features, models, and window sizes. In the features column, each number stands for total types of features used: 1 for TPN-r50 only; 4 for TPN-r50, TPN-r101, GTA, and irCSN; 5 for TPN-r101, VTN, GTA, irCSN, and I3D-Slow; 6 for TPN-r50, TPN-r101, VTN, GTA, irCSN, and I3D-Slow. In the spotting column, NV stands for NetVLAD++, and TF stands for Transformer. In the test column, ``-" means the result is not meaningful, due to the reason that test set is used in fine-tuning. In the challenge column, ``N/A" means it is not evaluated due to limited availability.}
    \begin{tabular}{@{}c|c|c|c|c|c@{}} 
        \toprule
        Row     &   Features                &   Spotting     &   Chunk/NMS  &    Test       &   Challenge\\
        \midrule
        1       &   ResNet                  &   NV           &   15/30      &   53.30       &   52.54   \\ 
        \midrule
        2       &   1                       &   NV           &   15/30      &   62.96       &   62.35   \\
        3       &   4                       &   NV           &   15/30      &   67.97       &   67.51   \\
        4       &   4 mega                  &   NV           &   15/30      &   -         &   67.57   \\
        5       &   4 mega                  &   NV ultra     &   15/30      &   -         &   68.68   \\
        6       &   5                       &   NV           &   15/30      &   72.64       &   71.08   \\
        7       &   5                       &   TF           &   7/20.      &   73.77       &   N/A     \\ 
        8       &   5                       &   NV           &   7/20       &   74.05       &   73.19   \\ 
        9       &   6 mega                  &   NV ultra     &   7/20       &   -         &   74.63   \\ 
        10      &   6 mega                  &   TF ultra     &   7/20       &   -         &   74.84   \\ 
        \bottomrule
    \end{tabular}
    \vspace{-0.1in}
\label{table:2}
\end{table}

\begin{table}[t]
    \small
    \centering
    \caption{Experimental results using different number of features; grounding models, including CALF-more-negative (Cmn), Transformer (TF); batch size (BS); and post-processing (PP) techniques, including filtering (FT), fusion (FS), and NMS. For features,  1  for  VTN;  2  for  TPN-r101 and  irCSN;  3 for TPN-r101, irCSN and GTA; 5  for  TPN-r101,  VTN,  GTA,  irCSN,  and  I3D-Slow. TF-u denotes training the transformer in ultra mode. The evaluation metric is average-AP as defined in \cite{Delige2020SoccerNetv2A}}
    \begin{tabular}{@{}c|c|c|c|c|c|c@{}} 
        \toprule
        Row     &   Features                &   Grounding     &  PP        &     BS           &   Challenge\\
        \midrule
        1       &   ResNet                  &   Cmn          &   N/A      &    32              &   40.75   \\ 
        \midrule
        2       &   2                       &   Cmn          &   N/A      &   32       &   52.11   \\
        3       &   3                       &   Cmn          &   N/A      &   64         &   55.79   \\
        4       &   1                       &   TF            &   N/A      &   128         &   58.62   \\ 
        5       &   5                       &   Cmn          &   N/A      &   128      &   59.26   \\
        6       &   5                       &   Cmn          &   FT       &128           &   62.19   \\
        7       &   5                       &   Cmn          &   FT,FS    &   128          &   64.01   \\ 
        8       &   5                       &   Cmn,TF-u      &   FT,FS,NMS&   128         &   71.59   \\ 
        9       &   5                       &   TF-u      &   FT       &   128         &   71.9    \\ 
        \bottomrule
    \end{tabular}
\label{table:results}
\end{table}




{\small
\bibliographystyle{ieee_fullname}
\bibliography{egbib}
}

\end{document}